\documentclass[11pt,a4paper]{article}
\usepackage[hyperref]{emnlp2020}
\usepackage{times}
\usepackage{latexsym}
\usepackage{amsmath,graphicx}
\usepackage{bm}
\usepackage{multirow}
\usepackage{url}
\usepackage{rotating}
\usepackage{pdflscape}
\usepackage{comment}
\usepackage{amssymb}
\usepackage{booktabs}
\usepackage{textcomp}
\usepackage{stfloats}

\usepackage{microtype}

\aclfinalcopy 

\title{An Empirical Study of Pre-trained Transformers for \\ Arabic Information Extraction}

\author{Wuwei Lan\textsuperscript{1}, Yang Chen\textsuperscript{2}, Wei Xu\textsuperscript{2}, Alan Ritter\textsuperscript{2} \\
\textsuperscript{1} Department of Computer Science and Engineering, Ohio State University\\
\textsuperscript{2} School of Interactive Computing, Georgia Institute of Technology\\
  {\small \tt lan.105@osu.edu   \{yang.chen, wei.xu, alan.ritter\}@cc.gatech.edu}
}

\date{}

\begin{document}
\maketitle
\begin{abstract}
Multilingual pre-trained Transformers, such as mBERT  \cite{devlin2018bert} and XLM-RoBERTa \cite{conneau2019unsupervised}, have been shown to enable the effective cross-lingual zero-shot transfer.  However, their performance on Arabic information extraction (IE) tasks is not very well studied. In this paper, we pre-train a customized bilingual BERT, dubbed GigaBERT, that is designed specifically for Arabic NLP and English-to-Arabic zero-shot transfer learning.  We study GigaBERT's effectiveness on zero-short transfer across four IE tasks: named entity recognition, part-of-speech tagging, argument role labeling, and relation extraction. Our best model significantly outperforms mBERT, XLM-RoBERTa, and AraBERT \cite{Antoun2020AraBERTTM} in both the supervised and zero-shot transfer settings. We have made our pre-trained models publicly available at \url{https://github.com/lanwuwei/GigaBERT}.


\end{abstract}

\section{Introduction}
Fine-tuning pre-trained Transformer models \cite{devlin2018bert,liu2019roberta,yang2019xlnet} has recently achieved state-of-the-art results on a wide range of NLP tasks where supervised training data is available.  When trained on multilingual corpora, BERT-based models have demonstrated the ability to learn multilingual representations that support zero-shot cross-lingual transfer learning surprisingly effectively \cite{wu2019beto,pires2019multilingual,lample2019cross}.

Without access to any parallel text or target language annotations, multilingual BERT \cite[mBERT;][]{devlin2018bert} even supports cross-lingual transfer for language pairs that are written in different scripts, for example, English-to-Arabic. However, the transfer learning performance still lags far behind where supervised data is available in the target language. In this paper, we explore to what extent it is possible to improve performance in the zero-shot scenario by building a customized bilingual BERT for English and Arabic, a particularly challenging language pair for cross-lingual transfer learning.  

We present GigaBERT, a customized BERT for English-to-Arabic cross-lingual transfer that is trained on newswire text in the Gigaword corpus \citep{graff2003english,parker2009arabic} in addition to Wikipedia and web crawl data.
We systematically compare our pre-trained models of different configurations against the mBERT \cite{devlin2018bert} and XLM-RoBERTa \cite[XLM-R;][]{conneau2019unsupervised}.  By using a customized vocabulary and code-switched data specifically created for English-to-Arabic transfer learning, our GiagBERT outperforms mBERT and \(\text{XLM-R}_\text{base}\) (both support more than 100 languages) on a range of IE tasks, including named entity recognition, part-of-speech tagging, argument role labeling, and relation extraction. Further performance gains are demonstrated by augmenting the pre-training corpus with synthetically generated code-switched data.  This demonstrates the usefulness of anchor points for zero-shot cross-lingual transfer learning. 
GigaBERT also performs well when annotated Arabic data is available, outperforming AraBERT \cite{Antoun2020AraBERTTM}, the state-of-the-art Arabic-specific BERT model, on various Arabic IE tasks.

\begin{table*}[!ht]
\tiny
\centering
\resizebox{\textwidth}{!}{%
\renewcommand{\arraystretch}{1.4}
\begin{tabular}{ l|lr|lrr|lc} 
 \toprule
\multirow{2}{*}{\bf Models} & \multicolumn{2}{c}{\bf Training Data} & \multicolumn{3}{c}{\bf Vocabulary} & \multicolumn{2}{c}{\bf Configuration} \\ 
& source & \#tokens (all/en/ar) & tokenization & size (all/en/ar) & cased & size & \#parameters \\
\midrule
 \(\text{AraBERT}\) & newswire &  2.5B/ -- /2.5B  & SentencePiece & 64k/ -- / 58k & no & base & 136M\\
 \(\text{mBERT}\) & Wiki &  21.9B/2.5B/153M  & WordPiece & 110k/53k/5k & yes & base  & 172M\\
  \(\text{XLM-R}_\text{base}\) & CommonCrawl &  295B/55.6B/2.9B  & SentencePiece & 250k/80k/14k & yes & base  & 270M\\
 \(\text{XLM-R}_\text{large}\) & CommonCrawl &  295B/55.6B/2.9B  & SentencePiece & 250k/80k/14k & yes & large  & 550M\\
 \(\text{GiagBERT-v0}\) & Gigaword & 4.7B/3.6B/1.1B  & SentencePiece & 50k/28k/19k & yes & base  & 125M\\
 \(\text{GigaBERT-v1}\) & Gigaword, Wiki & 7.4B/6.1B/1.3B & WordPiece & 50k/25k/23k & yes & base  & 125M\\
 \(\text{GigaBERT-v2/3}\) & Gigaword, Wiki, Oscar & 10.4B/6.1B/4.3B & WordPiece & 50k/21k/26k & no & base  & 125M \\
 \(\text{GigaBERT-v4}\) & Gigaword, Wiki, Oscar (+ code-switch) & 10.4B/6.1B/4.3B & WordPiece & 50k/21k/26k & no & base  & 125M \\
\bottomrule
\end{tabular}
}
\caption{\label{GigaBERT-versions} Configuration comparisons for AraBERT \cite{Antoun2020AraBERTTM}, mBERT \cite{devlin2018bert}, XLM-RoBERTa \cite{conneau2019unsupervised}, and GigaBERT (this work).}
\vspace{-.15cm}


\end{table*}

\section{Related Work}
The existing Arabic pre-trained models are either monolingual, such as hULMonA \cite{eljundi-etal-2019-hulmona} and AraBERT \cite{Antoun2020AraBERTTM}; or multilingual with several or over a hundred languages, such as mBERT \cite{devlin2018bert}, XLM \cite{lample2019cross}, and XLM-RoBERTa \cite{conneau2019unsupervised}. There is no bilingual pre-trained language model designed specifically for English-Arabic. \citet{KWMR20} pre-trained small-scale (e.g., 1GB data and 2M training steps) bilingual BERT for English-Hindi, English-Spanish, and English-Russian to study the impact of linguistic properties of the languages, the architecture of the model, and the learning objectives on cross-lingual transfer. \citet{kim2019qe} presented a bilingual BERT using multi-task learning for translation quality estimation with regards to English-Russian and English-German. \citet{wu2019emerging} focused on the bilingual XLM for English-French, English-Russian, and English-Chinese to analyze the cross-lingual transfer ability with domain similarity, anchor points, parameter sharing, and language similarity.

\section{GigaBERT}
We present five versions of GigaBERT pre-trained using the Transformer encoder \cite{vaswani2017attention} with \(\text{BERT}_\text{base}\) configurations: 12 attention layers, each has 12 attention heads and 768 hidden dimensions, which attributes 110M parameters. Table \ref{GigaBERT-versions} shows a detailed summary of the training data and model parameters.


\subsection{Training Data} 
We pre-train our GigaBERT models using the fifth edition of English and Arabic Gigaword corpora.\footnote{\url{https://catalog.ldc.upenn.edu/LDC2011T07} and \url{https://catalog.ldc.upenn.edu/LDC2011T11}} The Gigaword data consists of 13 million news articles\footnote{We flattened the Gigaword data with \url{https://github.com/nelson-liu/flatten_gigaword}.} and matches the domain of many NLP tasks. We split English and Arabic sentences without tokenization by a modified version of the Stanford CoreNLP tool \cite{manning2014stanford}.\footnote{In the early versions of GigaBERT (v0/1/2/3), we split Arabic sentences at period, exclamation, and question mark.}
We also add Wikipedia data processed by  WikiExtractor\footnote{\url{https://github.com/attardi/wikiextractor}} for better coverage. As the English Wikipedia (total 2.5B tokens) is much larger than the Arabic Wikipedia (total 0.15B tokens), we balance the pretraining data by (1) up-sampling the Arabic data by repeating the Wikipedia portion five times and the Gigaword portion three times; (2) adding the Arabic section of the Oscar corpus \cite{ortizsuarez:hal-02148693}, a large-scale multilingual dataset filtered from the Common Crawl.

\paragraph{Code-Switched Data Augmentation.} To further improve cross-lingual transfer capability, we leverage English-Arabic dictionaries to create synthetic code-switched training data \cite{conneau2019unsupervised}. We experimented with three dictionaries: PanLex \cite{kamholz-etal-2014-panlex}, MUSE \cite{conneau2017word}, and Wikipedia parallel titles. We extract parallel article titles in Wikipedia based on the inter-language links and the entities based on the Wikidata \cite{jiang2020neural}.\footnote{\url{https://github.com/clab/wikipedia-parallel-titles} and \url{https://dumps.wikimedia.org/wikidatawiki/entities/}} The dictionaries of PanLex, MUSE, Wikipedia contain 24K, 44K, 2M entries, respectively, and on overage 4.6, 1.4 and 1 translations per entry (English or Arabic). For training GigaBERT-v4, we code-switch up to 50\% random sentences for both English and Arabic and up to 30\% of tokens for each sentence. During the replacement process, we prioritize substitutions based on the Wikipedia titles, then PanLex and MUSE if the proportion of tokens being replaced has not reached 30\% for a given sentence. 


\subsection{Vocabulary}
The vocabulary size is critical to the performance of pre-training models, as it directly impacts the subword granularity and the number of parameters. The original English BERT \cite{devlin2018bert} uses a 30k vocabulary size for $\sim$3B tokens of training data, while the multilingual BERT and XLM-R have $\sim$5k and $\sim$14k Arabic subwords in their vocabularies respectively (Table \ref{GigaBERT-versions}).\footnote{We check the Unicode range of characters to classify word pieces as English or Arabic.} We choose a vocabulary size of 50k for our GigaBERT models based on preliminary experiments. For GigaBERT-v0, we use the unigram language model in the SentencePiece \cite{kudo2018sentencepiece} to create 30k cased English subwords and 20k Arabic subwords separately.\footnote{There are 633 word pieces shared by both languages. We add 633 unused symbols (e.g., unused-1, unused-2, etc.) to make up the 50k combined vocabulary.} For GigaBERT-v1/2/3/4, we did not distinguish Arabic and English subword units, instead, we train a unified 50k vocabulary using WordPiece \cite{wu2016google}.\footnote{We use Hugging Face's implementation: \url{https://github.com/huggingface/tokenizers}} The vocabulary is cased for GigaBERT-v1 and uncased for GigaBERT-v2/3/4, which use the same vocabulary.


\subsection{Optimization}
We use the official implementation of BERT \cite{devlin2018bert}  in TensorFlow for pre-training. We use Adam optimizer \cite{kingma2014adam} with a learning rate of 1e-4, \(\beta_1 = 0.9\), \(\beta_2 = 0.999\), L2 weight decay of 0.01. The learning rate is warmed up over the first 100,000 steps to a peak value of 1e-4, then linearly decayed. The dropout is set to 0.1 for all layers. We use the whole word mask for GigaBERT-v0 and the regular subword mask for v1/2/3/4. The batch size is set to 512. GigaBERT-v0/1/2 are trained for 1.2 million steps on Google Cloud TPUs with a max sequence length of 128. GigaBERT-v3 is additionally trained for 140k steps with a max sequence length of 512. The maximum number of masked LM predictions per sequence is set to 20 when max sequence length is 128 and set to 80 when max sequence length is 512. GigaBERT-v4 is trained from the GigaBERT-v3 checkpoint for another 140K steps on the code-switched data. We also experiment with different thresholds for the code-switched data augmentation, as well as training models from scratch on the code-switched data (Appendix \ref{code_switched_pretraining}).


\section{Experiments}

\subsection{Downstream IE Tasks}

We demonstrate the effectiveness of GigaBERT on named entity recognition (NER), part-of-speech tagging (POS), argument role labeling (ARL), and relation extraction (RE) tasks. We use the ACE 2005 corpus \cite{walker2006ace} in the NER, ARL, and RE evaluations, and use the Universal Dependencies Treebank v1.4 \cite{11234/1-1827} in the POS experiments. All of these datasets are from the news domain, as summarized in Table \ref{dataset statistics}. For NER, we use the same English document splits as~\citet{lu2015nner} and randomly shuffle Arabic documents into train/dev/test (80\%/10\%/10\%). For ARL and RE, we randomly shuffle both English and Arabic documents into train/dev/test (80\%/10\%/10\%). For POS, we follow the train/dev/test split by ~\citet{wu2019beto}. In the ARL fine-tuning experiment, we pair each trigger with its argument mentions as positive instances and with other entities in the sentence as negative instances. As for RE, we use gold relation mentions as positive examples and create negative examples by randomly pairing two entities in a sentence. We perform these tasks following the same fine-tuning pipeline as BERT \cite{devlin2018bert}. We feed input sentences into a pre-trained model, then extract the necessary hidden representations, i.e., all token representations for NER/POS and argument/entity spans for ARL/RE, before applying one linear layer for classification. We evaluate for each language in the standard supervised learning setting, as well as the zero-shot transfer learning setting from English to Arabic, where the model is trained on the annotated English training data and evaluated on the Arabic test set. 

\begin{table}[t!]
\centering
\resizebox{\columnwidth}{!}{
\renewcommand{\arraystretch}{1.4}
\begin{tabular}{lrrrr}
\toprule
\bf Task &  \bf \#Train (en/ar) & \bf \#Dev (en/ar) & \bf \#Test (en/ar) & \bf Metric\\ 
\hline
NER &  7634/2683 & 1005/322 & 1095/238 & \(\text{F}_\text{1}\) \\
POS & 12543/6174 & 2002/786 & 2077/704 & Acc \\
ARL & 21875/11587 & 3345/1221 & 2603/1568 & \(\text{F}_\text{1}\) \\
RE & 63177/32984 & 10218/4482 & 6861/4638 & \(\text{F}_\text{1}\) \\
\bottomrule
\end{tabular}
}

\caption{\label{dataset statistics} Statistics of the datasets for IE tasks.}
\vspace{-0.6cm}
\end{table}


\begin{table*}[!ht]
\small
\centering
\resizebox{\textwidth}{!}{%
\renewcommand{\arraystretch}{1.4}
\begin{tabular}{c |rrr|rrr|rrr|rrr } 
 \toprule
 \multirow{2}{*}{\bf Models} & \multicolumn{3}{c}{\bf NER (F$_1$)} & \multicolumn{3}{c}{\bf POS (Accuracy)} & \multicolumn{3}{c}{\bf ARL (F$_1$)}& \multicolumn{3}{c}{\bf RE (F$_1$)} \\ 
  & en & ar & en$\rightarrow$ ar & en & ar & en$\rightarrow$ ar & en & ar & en$\rightarrow$ ar & en & ar & en$\rightarrow$ ar \\ 
 \midrule
  \(\text{AraBERT}\) & - & {78.6} & - & - & {97.6} & - & - & {73.3} & - & - & {83.1} & - \\ 
  \(\text{mBERT}\)  & 80.3 & 72.9 & 30.8/31.1 & 97.0 & 97.3 & 50.8/50.8 & 70.4 & 64.5 & 44.4/45.9 & 77.9 & 75.3 & 30.1/30.1 \\ 
 \(\text{XLM-R}_\text{base}\)  & 81.0 & {81.5} & {43.5/43.5} & \textbf{97.8} & {97.6} & \underline{59.6/61.1} & 69.4 & 56.4 & \textbf{54.4}/\underline{53.7} & 78.2 & 79.2 & 40.4/36.0\\ 
  \(\text{GigaXLM-R}_\text{base}\)  & 82.0 & 80.8 & 45.4/45.0 & \underline{97.3} & \underline{97.7} & \textbf{60.7/61.4}  & 70.1 & 71.4 & {52.6/52.6} & 79.6 & 79.5 & \underline{43.3}/{44.0}\\ 
\(\text{GigaBERT-v0}\) & 79.1 & {76.6} & 43.9/45.9 & 96.8 & 97.5 & 49.7/54.1 & 69.1 & 66.1 & 42.3/42.2 & 76.6 & 72.5 & 21.5/20.9 \\ 
\(\text{GigaBERT-v1}\) & {82.8} & 72.9 & {49.1/49.1} & 97.2 & 96.6 & 51.9/52.2  & \textbf{72.8} & 67.7 & 44.6/45.5 & \textbf{80.4} & 73.2 & 36.0/31.1\\ 
\(\text{GigaBERT-v2}\) & {82.5} & 75.2 & {48.3/48.2} & {97.2} & \textbf{97.8} & 53.1/53.4 & 72.0 & 66.7 & 42.5/44.1 & 79.4 & 74.2 & 31.9/36.8\\
\(\text{GigaBERT-v3}\) & \underline{83.4} & \underline{83.1} & \underline{48.9/48.3} & {97.1} & \textbf{97.8} & {53.3/54.7} & \underline{72.3} & \textbf{76.5} & 51.0/51.0 & \underline{79.9} & \textbf{84.3} & \textbf{48.2}/\underline{46.8} \\ 
\(\text{GigaBERT-v4}\) & \textbf{83.8} & \textbf{84.1} & \textbf{51.5/51.5} & 97.1 & \underline{97.7} & 54.6/55.5 & 71.9 & \underline{73.9} & \underline{52.7}/\textbf{56.1} & 79.1 & \underline{83.6} & \underline{43.3}/\textbf{48.2} \\ 
 \midrule
\(\text{XLM-R}_\text{large}\) & {85.8} & {84.8}  & {49.3/50.4} & {98.0} & {97.8} & {61.7/61.2} & 72.3 & 73.4 & 58.0/57.4 & 83.2 & 82.1 & 52.5/57.5 \\ 
\(\text{GigaXLM-R}_\text{large}\) & 85.8 & 84.5 & 51.0/51.0 & 97.9 & 97.8 & 62.0/63.6 & 73.1 & 71.1 & 56.5/51.9 & 82.5 & 82.3 & 54.0/58.2 \\ 
 \bottomrule
\end{tabular}
}
\caption{\label{results_downstream} Evaluation on four Arabic IE tasks that compares AraBERT \cite{Antoun2020AraBERTTM}, multilingual BERT \cite{devlin2018bert}, XLM-RoBERTa \cite{conneau2019unsupervised}, GigaBERT/GigaXLM-R (this work). All models use \(\text{BERT}_\text{base}\) architecture except \(\text{XLM-R}_\text{large}\). GigaBERT-v4 is continued pre-training of GigaBERT-v3 on code-switched data. \(\text{GigaXLM-R}\) is domain adapted pre-training of \(\text{XLM-R}\) on Gigaword data.} 
\vspace{-0.3cm}
\end{table*}

\subsection{Implementations}

We implement the fine-tuning experiments with the PyTorch framework \cite{NEURIPS2019_9015} and choose hyperparameters by grid search.\footnote{The search range includes learning rate (1e-5. 2e-5. 5e-5. 1e-4), batch size (4, 8, 16, 32) and epoch number (3, 7, 10). } We set the learning rate to 2e-5, batch size to 8, max sequence length to 128, and the number of fine-tuning epochs to 7. Some exceptions include a learning rate of 1e-4 in NER experiments, max sequence length of 512, and batch size of 4 in RE experiments. For RE, we also use gradient accumulation to simulate the larger batch size of 32 when using models based on \(\text{BERT}_\text{large}\) architecture.


\subsection{Results and Analysis}
Table \ref{results_downstream} shows experimental results for the pre-trained models on both English and Arabic IE tasks. For the zero-shot transfer (en $\rightarrow$ ar), we report two scores on the Arabic test set, where the best checkpoint is selected based on the English dev set and the Arabic dev set, respectively. In summary, we find the key factors of improved pre-training performance are a large amount of training data in the target language, customized vocabulary, longer max length of sentence, and more anchor points from code-switched data. We also add experiments with \(\text{XLM-R}_\text{large}\) models as a reference, but the comparison focuses on the pre-trained models with \(\text{BERT}_\text{base}\) configuration for fairness.

\vspace{.1cm}
\noindent \textbf{Single-language Performance.} All versions of GigaBERT perform very competitively, especially the GigaBERT-v3/4. After adding Wikipedia and Oscar data, GigaBERT-v2 starts to outperform mBERT and \(\text{XLM-R}_\text{base}\) on most tasks. We find it crucial to continue training GigaBERT-v2 with a longer max sentence length of 512 word pieces, as the resulting GigaBERT-v3 model shows improvements in all four IE tasks. GigaBERT-v3 also outperforms AraBERT \cite{Antoun2020AraBERTTM}, the state-of-the-art Arabic-specific BERT model by a large margin, showing that our bilingual GigaBERT does not sacrifice per-language performance. It is worth noting that GigaBERT-v4 also has competitive single-language performance after training on the synthetically created code-switched data.



\vspace{.1cm}
\noindent \textbf{Cross-lingual Zero-shot Transfer Learning.} All pre-trained models show varied performance when we select checkpoints based on the English dev set and Arabic dev set, indicating that the best single-language performance does not necessarily imply the best cross-lingual performance. Compared to GigaBERT-v0, additional data used to train GigaBERT-v1/2 helps improve zero-shot transfer capability, even though the added data is not from the news domain. Different from previous works \cite{wu2019beto,pires2019multilingual} that attribute cross-lingual ability to shared subwords, GigaBERT-v3 has nearly no shared word pieces or scripts between English and Arabic, but still shows strong cross-lingual performance. We hypothesize the Transformer encoder projects similar contextual representations and enables cross-lingual transfer \cite{wu2019emerging}.


\vspace{.1cm}
\noindent \textbf{Code-Switched Pre-training.} We show that we can further improve GigaBERT's cross-lingual transfer capability with a carefully designed code-switching procedure. Our GigaBERT-v4 pre-trained with code-switched data shows significant improvement over GigaBERT-v3, achieving new state-of-the-art for zero-shot transfer from English to Arabic on NER, ARL, and RE. Our code-switched pre-training differs from ~\citet{wu2019emerging} in two aspects: 1) we explored multiple bilingual dictionaries, including PanLex \cite{kamholz-etal-2014-panlex}, MUSE \cite{conneau2017word} and Wikipedia titles, while MUSE appears to be the most effective; 2) we keep at least half of the sentences unchanged to balance between real data and artificial data. In practice, the generated data for GigaBERT-v4 has 47.4\% of the sentences code-switched. We present more comparison experiments using varied code-switching mixes and different bilingual lexicons in Appendix \ref{code_switched_pretraining}.


\vspace{.1cm}
\noindent \textbf{Domain-adapted Pre-training.} We also explore whether XLM-RoBERTa can be improved by additional pre-training on Gigaword data, as ~\citet{gururangan-etal-2020-dont} have shown that the continued pre-training with in-domain data is helpful. 
We create GigaXLM-R models by continuing pre-training from \(\text{XLM-R}_\text{base}\) and \(\text{XLM-R}_\text{large}\) checkpoints in the Fairseq toolkit \cite{ott2019fairseq} for 500k steps on shuffled Arabic and English Gigaword corpus (max sequence length 512 and batch size 4). Although only $\sim$1\% of the Gigaword corpus is used in this continued training step due to computing resource limit, GigaXLM-R still improves zero-shot transfer performance for NER, POS, and RE over the original \(\text{XLM-R}\) models as shown in Table \ref{results_downstream}. We could expect more performance improvement with a larger batch size and longer training time.


\begin{figure}
    \centering
    \includegraphics[width=.95\linewidth]{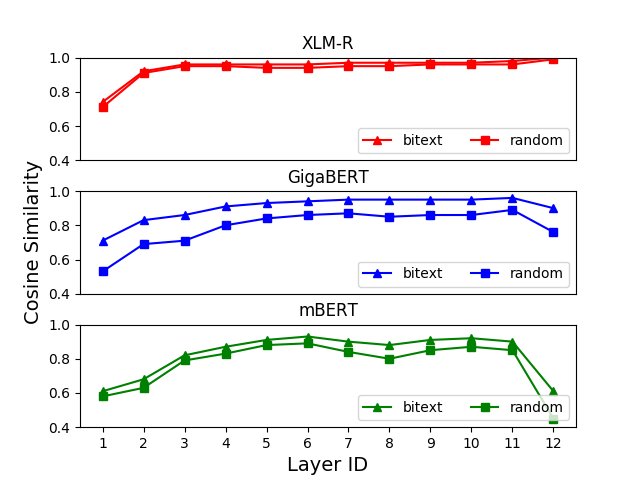}
    \caption{Cosine similarity between sentence representations of parallel sentences (bitext) and randomly paired sentences (random).}
    \vspace{-0.6cm}
    \label{fig:analysis}
\end{figure}

\vspace{.1cm}
\noindent 
\textbf{Embedding Space Analysis.} We further analyze the semantic similarity of parallel English-Arabic sentence representations and find that GigaBERT is able to distinguish parallel sentences from randomly paired sentences more effectively compared to its counterparts. Our hypothesis is that cross-lingual representations for parallel English-Arabic sentences should be similar, but randomly paired sentences should be dissimilar. To evaluate cross-lingual similarity, we extract sentence representation of 5340 English-Arabic parallel sentences from the GALE corpus\footnote{\url{https://catalog.ldc.upenn.edu/LDC2014T10}} and the same number of randomly paired sentences with pre-trained models across all 12 layers. We use the average of hidden representations, excluding [CLS] and [SEP], as a sentence representation. Cosine similarity is calculated for each sentence pairs and averaged across the whole corpus. In Figure~\ref{fig:analysis}, GigaBERT shows high similarity between parallel sentences and low similarity between randomly paired sentences. A clear separation for two types of paired sentences is shown across all the layers. In contrast, XLM-R is not able to distinguish between them but shows high similarity scores. mBERT shows low similarity in both cases. This suggests that our GigaBERT preserves language independent semantic information in the sentence representations, which might contribute to the competitive performance in downstream IE tasks.


\section{Conclusions}
In this paper, we show that the performance of zero-shot cross-lingual transfer can be improved by training customized bilingual BERT for a given language pair and text domain.  We pre-trained several masked language models (GigaBERTs) for Arabic-English and conducted a focused study on information extraction tasks in the newswire domain. The experiments show that our GigaBERT model outperforms multilingual BERT, XLM-RoBERTa, and the monolingual AraBERT on NER, POS, ARL and RE tasks. We also achieve the new state-of-the-art performance fro zero-shot transfer learning from English to Arabic. We additionally studied code-switched pre-training for GigaBERT and domain-adapted pre-training for XLM-RoBERTa. 

\section*{Acknowledgement}
We thank Nizar Habash and anonymous reviewers for their valuable suggestions. We also thank the Google TFRC program for providing free TPU access. This material is based in part on research sponsored by the NSF (IIS-1845670), ODNI, and IARPA via the BETTER program (2019-19051600004), DARPA via the ARO (W911NF-17-C-0095) in addition to an Amazon Research Award. The views and conclusions contained herein are those of the authors and should not be interpreted as necessarily representing the official policies, either expressed or implied, of ODNI, IARPA, ARO, DARPA, or the U.S. Government. The U.S. Government is authorized to reproduce and distribute reprints for governmental purposes notwithstanding any copyright annotation therein.

\bibliography{emnlp2020}
\bibliographystyle{acl_natbib}

\clearpage

\appendix

\section{Comparison Experiments for Code-Switched Pre-training}
\label{code_switched_pretraining}

\begin{table*}[hb!]
\small
\centering
\resizebox{\textwidth}{!}{%
\renewcommand{\arraystretch}{1.4}
\begin{tabular}{c |ccc|ccc|ccc|ccc } 
 \toprule
 \multirow{2}{*}{\bf Models} & \multicolumn{3}{c}{\bf NER (F$_1$)} & \multicolumn{3}{c}{\bf POS (Accuracy)} & \multicolumn{3}{c}{\bf ARL (F$_1$)}& \multicolumn{3}{c}{\bf RE (F$_1$)} \\ 
  & en & ar & en$\rightarrow$ ar & en & ar & en$\rightarrow$ ar & en & ar & en$\rightarrow$ ar & en & ar & en$\rightarrow$ ar \\ 
 \midrule
\(\text{s1-0.5-0.3-all}\) & 82.1 & 83.3 & 49.7 & 97.0 & 97.7 & 58.3 & 72.4 & 74.4 & 48.6 & 80.0 & 84.1 & 47.0 \\ 
\(\text{s1-1.0-0.5-all}\) & 83.1 & 82.9 & 48.3 & 97.0 & 97.7 & 55.0 & 71.1 & 74.6 & 46.9 & 79.0 & 82.7 & 40.2 \\ 
\(\text{s1-0.5-0.3-pm}\) & 83.5 & 83.9 & 51.3 & 97.2 & 97.8 & 56.9 & 71.4 & 73.4 & 38.3 & 74.9 & 82.8 & 47.6 \\ 
\(\text{s1-0.5-0.3-m}\) & 82.4 & 84.7 & 52.9 & 97.1 & 97.8 & 58.6 & 70.7 & 72.4 & 52.1 & 77.3 & 83.7 & 46.0 \\ 
\(\text{s1-0.5-0.1-mw}\) & 83.1 & 83.9 & 52.2 & 97.2 & 97.6 & 55.0 & 71.7 & 72.7 & 49.0 & 78.2 & 84.1 & 54.0 \\ 
\(\text{s1-0.5-0.3-mw}\) & 83.3 & 83.3 & 53.5 & 97.1 & 97.7 & 56.0 & 71.9 & 72.8 & 46.8 & 79.2 & 84.2 & 44.3 \\
\(\text{s1-1.0-0.3-mw}\) & 82.7 & 84.4 & 48.2 & 97.1 & 97.7 & 56.1 & 70.6 & 72.7 & 51.4 & 77.9 & 84.6 & 47.3 \\
\(\text{s1-1.0-0.001-mw}\) & 83.4 & 83.8 & 54.1 & 97.2 & 97.7 & 55.1 & 72.3 & 73.3 & 48.0 & 78.7 & 83.5 & 41.2 \\
\(\text{s1-0.5-0.3-w}\) & 82.8 & 83.8 & 49.9 & 97.1 & 97.8 & 53.8 & 71.4 & 73.8 & 50.8 & 77.1 & 82.7 & 54.3 \\
\(\text{s2-0.5-0.3-all}\) & 83.8 & 84.1 & 51.5 & 97.1 & 97.7 & 55.5 & 71.9 & 73.9 & 56.1 & 79.1 & 83.6 & 48.2 \\ 
\(\text{s2-1.0-0.5-all}\) & 82.2 & 83.7 & 51.7 & 97.0 & 97.8 & 56.1 & 71.3 & 74.5 & 51.1 & 79.2 & 82.0 & 45.8 \\ 
\(\text{s2-0.5-0.3-pm}\) & 83.2 & 83.8 & 50.9 & 97.1 & 97.7 & 55.7 & 72.0 & 73.7 & 48.4 & 79.3 & 82.9 & 45.3 \\ 
\(\text{s2-0.5-0.3-m}\) & 83.4 & 83.4 & 52.9 & 97.2 & 97.7 & 52.9 & 71.0 & 73.9 & 55.0 & 78.8 & 83.5 & 52.5 \\ 
\(\text{s2-0.5-0.1-mw}\) & 83.0 & 85.1 & 52.7 & 97.2 & 97.8 & 53.6 & 71.9 & 75.0 & 50.0 & 79.0 & 83.7 & 52.2 \\ 
\(\text{s2-0.5-0.3-mw}\) & 83.4 & 85.0 & 51.0 & 97.1 & 97.7 & 52.4 & 72.2 & 74.9 & 49.3 & 81.0 & 83.7 & 49.8 \\
\(\text{s2-1.0-0.3-mw}\) & 83.2 & 83.7 & 50.2 & 97.0 & 97.7 & 53.5 & 71.0 & 71.8 & 54.7 & 67.2 & 81.3 & 42.9\\
\(\text{s2-1.0-0.001-mw}\) & 83.6 & 84.2 & 49.6 & 97.4 & 97.7 & 52.3 & 71.8 & 73.2 & 51.2 & 79.0 & 84.0 & 42.6 \\
\(\text{s2-0.5-0.3-w}\) & 83.7 & 83.9 & 50.4 & 97.2 & 97.7 & 53.1 & 72.6 & 74.4 & 48.2 & 76.2 & 83.6 & 47.4 \\
 \bottomrule
\end{tabular}
}
\caption{\label{results_code_switching} Comparison experiments of different code-switching configurations. The model name is composed of four parts: {\textbf{s1} (pre-train from scratch)/ \textbf{s2} (continue pre-training)}, {sentence replacement threshold}, {token replacement threshold} and {bilingual lexicons}, where \textbf{all} uses PanLex, MUSE and Wiki titles, \textbf{pm} uses PanLex and MUSE, \textbf{mw} uses MUSE and Wiki, \textbf{m} uses MUSE only and \textbf{w} uses Wiki only. The model \text{s2-0.5-0.3-all} is GigaBERT-v4 in the paper. The best checkpoint for en$\rightarrow$ ar is selected with Arabic dev set.} 
\end{table*}

Given the English and Arabic monolingual corpus and the bilingual lexicons, we have different thresholds to control the code-switched data generation: 1) the percentage of sentences being code-switched within the whole corpus, we set sentence replacement threshold to limit the changed sentences; 2) the percentage of tokens being replaced within the sentence, we set token replacement threshold to limit the changed tokens; 3) the choice of bilingual lexicons, where we explore different combinations of PanLex, MUSE and Wiki titles. With the generated code-switched data, we can pre-train GigaBERT from scratch or load the existing checkpoint (GigaBERT-v3) for continued pre-training, which are s1 and s2 in Table \ref{results_code_switching}, respectively. 

As shown in Table \ref{results_code_switching}, it's better to keep some sentences unchanged for code-switched pre-training. The continued pre-training (s2) shows slightly better performance than that training from scratch (s1). During the data augmentation, we need to keep a relatively low ratio for token replacement. The results also reveal that the MUSE dictionary is very promising, which outperforms the combinations of all dictionaries in some cases.

\end{document}